\documentclass{article}

\usepackage{PRIMEarxiv}
\usepackage[T1]{fontenc}
\usepackage{hyperref}
\usepackage{url}
\usepackage{booktabs}
\usepackage{amsfonts}
\usepackage{nicefrac}
\usepackage{microtype}
\usepackage{lipsum}
\usepackage{fancyhdr}
\usepackage{graphicx}
\usepackage{amsmath}
\usepackage{adjustbox}
\usepackage{booktabs}
\usepackage{tabularx}
\usepackage{array}
\usepackage{multirow}
\usepackage{graphicx}

\DeclareMathOperator*{\argmax}{arg\,max}

\graphicspath{{media/}}

\title{Enhanced Textual Feature Extraction for Visual Question Answering: A Simple Convolutional Approach}

\author{
  Zhilin Zhang \\
  Tandon School of Engineering \\
  New York University \\
  New York, USA \\
  \texttt{zz10068@nyu.edu} \\
\And
  Fangyu Wu \\ 
  Department of Computer Science \\ 
  University of Illinois Urbana-Champaign \\ 
  Urbana, USA \\ 
  \texttt{fangyuw2@illinois.edu} \\
}

\begin{document}
\maketitle

\begin{abstract}
Visual Question Answering (VQA) has emerged as a highly engaging field in recent years, with increasing research focused on enhancing VQA accuracy through advanced models such as Transformers. Despite this growing interest, limited work has examined the comparative effectiveness of textual encoders in VQA, particularly considering model complexity and computational efficiency. In this work, we conduct a comprehensive comparison between complex textual models that leverage long-range dependencies and simpler models focusing on local textual features within a well-established VQA framework. Our findings reveal that employing complex textual encoders is not invariably the optimal approach for the VQA-v2 dataset. Motivated by this insight, we propose ConvGRU, a model that incorporates convolutional layers to improve text feature representation without substantially increasing model complexity. Tested on the VQA-v2 dataset, ConvGRU demonstrates a modest yet consistent improvement over baselines for question types such as Number and Count, which highlights the potential of lightweight architectures for VQA tasks, especially when computational resources are limited.
\end{abstract}


\section{Introduction}
Visual Question Answering (VQA) \cite{antol2015vqa} has emerged as an increasingly interesting area of research at the intersection of computer vision and natural language processing. The objective of VQA is to answer questions about a given image, requiring models capable of understanding both visual and textual modalities. One particular task in the VQA domain that attracts additional attention is the counting task. This specific task requires the model to quantitatively identify the number of objects in a given image. Despite its apparent simplicity, this task demands precise identification, localization, and counting of objects, and avoiding potential pitfalls like multiple counting or not acknowledging occluded items. 

Advanced sequential models, such as Transformers, have seen great success in various tasks and have been widely adopted. However, these models are not always the best choice for every task, particularly in scenarios where balancing model performance with computational cost is essential. As highlighted in previous studies that evaluate models under specific conditions \cite{du2024comparison}, carefully considering model selection and exploring simple yet effective improvements is also crucial for optimizing task performance.

In this paper, we pose the question: Are complex sequential models always the most suitable approach for handling textual modality in VQA tasks? To investigate this problem, we conduct comprehensive experiments applying both complex models like Transformer Encoder and attention-based models, as well as simpler structures such as RNNs and CNNs on a well-established VQA-v2 dataset. We focus our analysis on different text feature extraction methods and their impact on model accuracy. We discover that simpler models, specifically those good at capturing local interdependencies within the text, can provide improvements over some complex models. Furthermore, the incorporation of convolutional layers into simpler structures like Gated Recurrent Units (GRU), forming what we term as ConvGRU, proves quite feasible.

Our main contributions are threefold: (1) We present ConvGRU, a GRU-based text encoder enhanced by applying convolutional layers. Experimental results demonstrate that leveraging local textual features with ConvGRU can marginally improve VQA accuracy without increasing computational costs. (2) We conduct an extensive comparison of various VQA text encoders, and find that complex models such as Transformer Encoders may underperform compared to simpler architectures better at extracting local textual features.

\begin{figure}
  \centering
    \includegraphics[width=\textwidth]{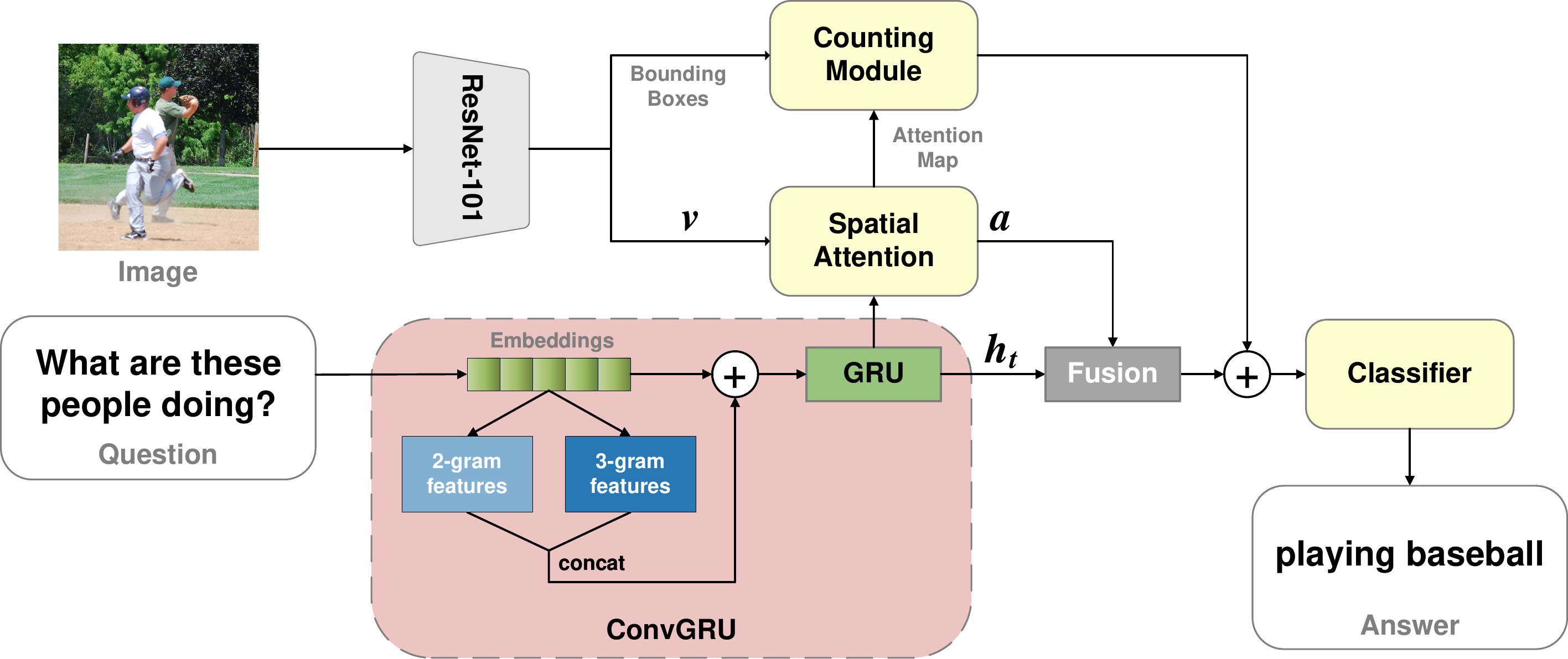}
  \caption{Overview of the VQA architecture highlighting the ConvGRU module. Questions are embedded and processed through convolutional layers to capture local semantics, then integrated via a GRU unit. Image features are extracted using ResNet-101 with spatial attention focusing on key areas. The fusion module combines image and text information, and the classifier outputs the answer. Counting, attention and classifier modules follow the configurations of.} 
  \label{fig:overall}
\end{figure}

\section{Related Work}

\textbf{Visual Question Answering.}  Visual Question Answering (VQA) has rapidly evolved since \cite{antol2015vqa} first introduced the concept of free-form and open-ended VQA tasks, accompanied by a benchmark dataset and evaluation metrics. \cite{goyal2017making} introduces complementary image-question pairs to mitigate the imbalance caused by language priors. Early VQA research primarily utilized Convolutional Neural Networks (CNNs) and Recurrent Neural Networks (RNNs) for image and question encoding, respectively \cite{antol2015vqa, malinowski2015ask, wang2018learning, liu2018ivqa}. \cite{malinowski2015ask} exemplifies this approach through a two-branch neural network combining a CNN-based image encoder with an LSTM question encoder. The emergence of Transformer models has then catalyzed the development of cross-modal architectures in the VQA domain. Notably, models like ViLBERT \cite{lu2019vilbert}, VisualBERT \cite{li2019visualbert}, and CLIP \cite{radford2021learning} have led the transition towards unified frameworks. These models have been instrumental in recent VQA research efforts \cite{garderes2020conceptbert, ravi2023vlc, song2022clip, zhou2020unified}, showcasing remarkable results on VQA tasks. 

\textbf{Spatial Attention in VQA.}  Spatial attention mechanism is adopted in order to improve the performance of VQA and mimic human behavior, i.e., looking at certain parts of the regions \cite{yang2016stacked, xu2016ask, seo2016bidirectional}. Stacked attention networks (SAN) \cite{yang2016stacked} and Spatial Memory Network (SMem) \cite{xu2016ask} similarly utilize two-stage attention framework to produce better glimpses through multiple reasoning. To generate a query-aware context representation without early summarization, \cite{seo2016bidirectional} proposes a multi-stage hierarchical bi-directional attention network.

\textbf{Text Representations in VQA.} The task of Visual Question Answering (VQA) necessitates effective textual representation to comprehend questions that guide the visual understanding. Traditional approaches have predominantly employed Recurrent Neural Networks (RNNs), including Gated Recurrent Units (GRU) and Long Short-Term Memory (LSTM) networks \cite{antol2015vqa, malinowski2015ask, kazemi2017show}. Their ability to model the temporal dependencies inherent in natural language questions makes them robust baselines for text representation in VQA tasks. Beyond RNNs, the Text Convolutional Neural Network (TextCNN) \cite{kim-2014-convolutional} emerges as an alternative that leverages convolutional operations to extract local features, offering a different perspective on capturing textual semantics \cite{wang2018learning}. The advent of the Transformer \cite{vaswani2017attention} has introduced a shift towards attention-based structures for text representation. The Transformer Encoder, with its self-attention mechanism, allows for the direct modeling of relationships between all words in a question, irrespective of their positional distances. Building on the success of Transformers, BERT (Bidirectional Encoder Representations from Transformers) \cite{devlin2018bert} has emerged as another milestone in text representation adept at grasping the nuanced context of words in a sentence from both directions. This progression has culminated in Transformers and BERT becoming the dominant models in recent VQA works \cite{garderes2020conceptbert, yang2021tap}.

\section{Methodology}

\subsection{Problem Formulation}
The primary objective of Visual Question Answering (VQA) is to accurately predict an answer given an image and a corresponding natural-language question. Therefore, the VQA task can be regarded as a classification problem, and its goal is to identify the most probable answer $\hat{a}$ from a predefined set of possible answers, based on the provided image $i$ and a question $q$. Formally, this is expressed as: 
\begin{equation}
\hat{a} = \argmax_a P(a|i,q)
\end{equation}

where $a$ denotes a potential answer within the set $A=\{a_1,a_2,a_3,...,a_n\}$, with $n$ indicating the total number of possible answers. In scenarios involving counting tasks within VQA, $\hat{a}$ signifies the number that most accurately responds to the posed question concerning the image.

\subsection{Method Overview}
Our methodology seeks to improve VQA accuracies by focusing on enhanced text processing techniques, leveraging the foundational work of \cite{zhang2018vqacount}. We introduce a model that primarily innovates in textual feature extraction through a novel Convolutional GRU framework, and also integrate existing methods for image feature extraction and a counting component to address both visual data and numerical queries effectively.

At the heart of our proposed model, as depicted in Figure \ref{fig:overall}, is the integration of a Convolutional GRU framework for improved extraction of textual features from questions. For image modality, we employ a pre-trained R-CNN model for extracting features, accompanied by a stacked attention mechanism and a counting module from the base model.

\subsection{Standard Models}
We directly adopt several established methods to highlight our contributions in the textual modality and give a fair comparison of different textual encoders under the same setting.

\subsubsection{Image Encoder} \label{Image Encoder} 
We employ a pre-trained R-CNN model with a ResNet-101 backbone \cite{he2016deep} for image feature extraction, leveraging its architecture optimized for depth and computational efficiency through bottleneck blocks. Table \ref{tab:resnet} provides an overview of ResNet-101 structure and the arrangement of its bottleneck blocks.

\begin{table}
 \caption{Detailed Structure of ResNet-101 Used for Image Feature Extraction. The table outlines the configuration of layers in ResNet-101, including convolutional layers, max pooling, and the bottleneck blocks.}
  \centering
  \begin{adjustbox}{width=\textwidth,center}
  \begin{tabular}{c|cccccc}
    \toprule
    \textbf{Layer name} & Conv1 & Max Pooling & Conv2\_x & Conv3\_x & Conv4\_x & Conv5\_x \\
    \midrule
    \textbf{Structures} & 7×7, 64 & 3×3 & $ \left(\begin{array}{c}1\times1, 64 \\ 3\times3, 64 \\ 1\times1, 256\end{array}\right)$×3 & $\left(\begin{array}{c}1\times1, 128 \\ 3\times3, 128 \\ 1\times1, 512\end{array}\right)$×4 & $\left(\begin{array}{c}1\times1, 256 \\ 3\times3, 256 \\ 1\times1, 1024\end{array}\right)$×23 & $\left(\begin{array}{c}1\times1, 512 \\ 3\times3, 512 \\ 1\times1, 2048\end{array}\right)$×3 \\
    \bottomrule
  \end{tabular}
  \end{adjustbox}
  \label{tab:resnet}
\end{table}

\subsubsection{Stacked Attention Mechanism}
We utilize the stacked attention network proposed by \cite{yang2016stacked}, which employs multiple layers of attention to pinpoint relevant areas within an image iteratively.

At each step $t$, the mechanism calculates a glimpse $x_t$, a weighted mix of image features $v$ and question features $q$, using:
\begin{equation}
x_t = \sigma(\text{Conv}([\text{dropout}(v); \text{Tile}(\text{dropout}(q))]))
\end{equation}

Here, $\sigma$ represents the softmax function, applying the convolutional operation $Conv$ on the combined and processed features, where $dropout$ enhances model generalization, and $Tile$ replicates $q$ across all spatial dimensions of $v$.

The attention map $x_t$ is then employed to calculate a focused weighted sum of image features:
\begin{equation}
a_t = \sum_{i} x_t^i \cdot v^i  
\end{equation}

where $x_t^i$ and $v^i$ correspond to the $i$-th elements of $x_t$ and $v$, respectively.

By applying this attention sequence iteratively, the network refines its focus, progressively integrating pertinent image information to inform the answer generation process.

\subsubsection{Counting Module}
Adopted from \cite{zhang2018vqacount}, the counting module integrates attention weights $a$ and bounding boxes $b$ as input to enable precise object counting. The process unfolds through several key operations, streamlined for clarity:

\begin{enumerate}
    \item \textbf{Attention Matrix Formation}: Attention weights are transformed into an attention matrix $A = aa^T$, mapping the relationship between different object proposals.
    \item \textbf{Duplicate Removal}: Intra-object duplicates are eliminated by applying a mask created from the inverse of the Intersection over Union (IoU) scores between bounding boxes: 
    \begin{equation}
        D_{ij} = 1 - IoU(b_i, b_j)
    \end{equation}
    
    This produces a refined attention matrix $\Tilde{A} = A \odot D$, where $\odot$ denotes element-wise multiplication.
    \item \textbf{Inter-Object Differentiation}: Inter-object duplicates are addressed by calculating a uniqueness score for each proposal. This involves assessing the similarity between proposals based on their attention weights and adjusted attention matrix, $\Tilde{A}$. Each proposal's score is inversely related to the number of similar proposals.
    \item \textbf{Count Matrix and Output}: The final counting matrix $C$ is computed by considering the scaled attention matrix and adding self-loops based on the uniqueness scores. The count output vector $O$ is derived from $C$, adjusted for the degree of overlap and distance from expected counts.
\end{enumerate}

This module's design enables the VQA model to differentiate between and accurately count overlapping objects, thereby enhancing the model's overall counting accuracy.

\subsubsection{Fusion and Classifier Module}
Following the methodology described in \cite{zhang2018vqacount}, the fusion of features is executed as follows:
\begin{equation}
Fuse_{out} = ReLU(w_1 x+w_2 y)-(w_1 x-w_2 y)^2
\end{equation}

In this equation, $w_{1}$ and $w_{2}$ are weights assigned to features $x$ (visual) and $y$ (textual), respectively. This method combines linear and quadratic interactions, activated by $ReLU$, to merge these feature sets effectively.

For classification, a multi-modal approach is utilized, concatenating visual and question representations, along with the counting module's output. This concatenated result undergoes $ReLU$ activation and is further refined through batch normalization before proceeding to a fully-connected layer designed for a 3000-category classification task.

The model's accuracy is assessed using the Negative Log-Likelihood Loss Function, defined as:
\begin{equation}
Loss=-\frac{1}{N} \sum_{n=1}^{N}\log P(a_{n}|i,q)
\end{equation}

Here, the loss is averaged across all correct answers to compute the final value, ensuring a comprehensive evaluation of the model's performance.

\subsection{Convolutional GRU for Text Encoding} \label{CGRU}
In Visual Question Answering (VQA), textual feature extraction traditionally utilizes Recurrent Neural Networks (RNNs) like Long Short-Term Memory (LSTM) and Gated Recurrent Units (GRU) due to their proficiency in modeling sequential data. GRUs are often favored for their computational efficiency, especially given the concise and straightforward nature of questions in VQA tasks. Our main contribution lies in enhancing the GRU architecture by incorporating convolutionally extracted n-gram features as inputs instead of direct word embeddings. This approach is theoretically grounded in the understanding that local n-gram patterns are crucial for capturing semantic meaning in text \cite{wang2016semantic}.

We propose a one-dimensional convolutional approach as a more effective alternative for feature extraction. By employing kernels of sizes 2 and 3, we capture bi-gram and tri-gram features, respectively, which are then processed through distinct padding strategies-asymmetrical for bi-grams to prioritize the first word of the sequence, and symmetrical for tri-grams to preserve context evenly. Similar to the observations in \cite{wu2019convolution} regarding the shift problem encountered with even-numbered convolutions in image processing, we identify a parallel issue in the realm of one-dimensional text convolutions. To avoid the potential complexity and overfitting that might arise from directly addressing this shift problem, we choose to simply add a unit of padding at the beginning of sequences to accurately identify the initial words of a question, which are crucial for defining its category. For instance, ``what'' typically signals a query about an object, while ``what color'' suggests a question about color. The utility of this asymmetric head padding technique is also validated by the results in Section \ref{sec:results}.

\begin{figure}
  \centering
  \includegraphics[width=\textwidth]{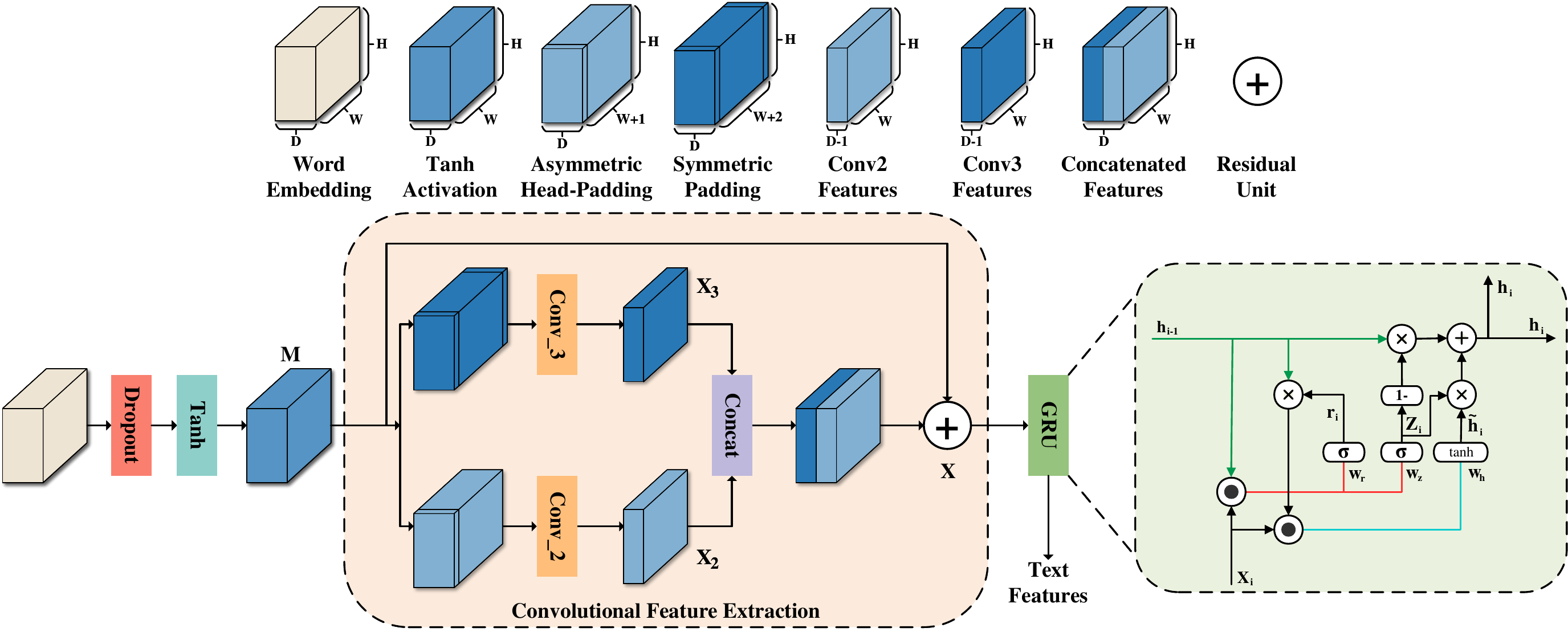}
  \caption{Schematic representation of the textual modality feature extraction model. Word embeddings are processed through dropout, Tanh activation, padding, and convolutional layers to capture multi-scale features. A residual unit is applied to ensure even questions with subtle n-gram features retain robust initial representations. Then, features are concatenated and fed into a GRU to extract sequential semantics. The inset details the GRU's internal gating mechanisms.}
  \label{fig:CGRU-23concat}
\end{figure}

Figure \ref{fig:CGRU-23concat} outlines our Convolutional GRU (ConvGRU) model's architecture, highlighting the integration of convolutional n-gram feature extraction with GRU processing. Given a sequence of question words $q$, each word is first embedded into a high-dimensional space, yielding the matrix $E \in \mathbb{R}^{B \times L \times d} $, where $B$ is the batch size, $L$ is the sequence length, and $d$ is the embedding size. Following a dropout and Tanh activation, the matrix is calculated as:
\begin{equation}
M = Tanh(Dropout(E))
\end{equation}

Convolution operations with kernels of sizes 3 and 2 are then applied to $M$, and their outputs are concatenated along with a residual connection to $M$:
\begin{equation}
X_{3} = \text{Conv3(SymPad(M))} \\
\end{equation}
\begin{equation}
X_{2} = \text{Conv2(AsyHeadPad(M))} \\
\end{equation}
\begin{equation}
X = [ X_{3};X_{2} ] + M \\
\end{equation}

where $SymPad$ represents symmetric padding, and $AsyHeadPad$ represents asymmetric head-padding.

The GRU then processes the enriched feature set $X$, calculating the update gate $z_{i}$, reset gate $r_{i}$, and the new memory cell $\Tilde{h_{i}}$ at each time step $i$, leading to the updated hidden state $h_{i}$:
\begin{equation}
r_{i} = \sigma (W_{r} \cdot x_{i} + U_{r} \cdot h_{i-1} + b_{r}) \\
\end{equation}
\begin{equation}
z_{i} = \sigma (W_{z} \cdot x_{i} + U_{z} \cdot h_{i-1} + b_{z}) \\
\end{equation}
\begin{equation}
\Tilde{h_{i}} = \text{Tanh} (W_{h} \cdot x_{i} + U_{h} \cdot (r_{i} \odot h_{i-1}) + b_{h}) \\
\end{equation}
\begin{equation}
{h_{i}} = (1-z_{i}) \odot h_{i-1} + z_{i} \odot \Tilde{h_{i}} \\
\end{equation}

where $\sigma$ is the sigmoid function, $\odot$ is the element-wise multiplication, $W_{r}, W_{z}, W_{h}, U_{r}, U_{z}, U_{h}$ are weight matrices, and $b_{r}, b_{z}, b_{h}$ are bias vectors. 

With this schema, our Convolutional GRU improves upon traditional methods by adeptly capturing and emphasizing pivotal textual features, thus providing a robust foundation for more accurate responses.

\section{Experiments}

\subsection{Dataset and Evaluation Metric}
\textbf{VQA-v2 Dataset.}  A notable change of Visual Question Answering  was the transition from the VQA dataset to VQA-v2 dataset in order to address a number of biases and language priors found in the original dataset \cite{goyal2017making}. For instance, ``blue'' is frequently selected as the answer to questions beginning with ``What color is ...'' regardless of the image content, which may lead to improper correlations between the question text and the answer. To cope with these constraints, VQA-v2 introduces complementary pairs, wherein there is a ``complementary'' image that yields a different answer to the same question for every question linked to an image. In our experiments, VQA-v2 dataset is chosen as our primary focused task, and we assess our model on validation set after training on the training set.

\textbf{Evaluation metric}. In the realm of VQA, accuracy is calculated using a unique method. For each question, the model's answer is compared with human responses. The accuracy for a particular answer is calculated by:
\begin{equation}
    Acc=\frac{1}{|A|} \sum_{a \in A,|a|=|A|-1}\min(\frac{1}{3}Agree(a),1)
    \label{eq:acc}
\end{equation}

where $A$ is the human answer set including 10 answers, and $Agree(\cdot)$ is the number of human answers same with the predicted one. An answer is considered correct if at least 3 humans give the same answer to ensure accuracy while accounting for the possibility of multiple correct answers to the same question.

To address language priors and answer imbalance in the dataset, we also employ the \textbf{Balanced Pairs Accuracy} \cite{teney2018tips} metric. It forms complementary pairs by asking the same question about two different images requiring different answers. The model receives credit only if it answers both correctly. Mathematically, Balanced Pairs Accuracy is calculated as:
\begin{equation}
    \overline{Acc}=\frac{1}{B}\sum_{k=1}^{B}\delta(Acc_{k}^{(1)} == 1 \,and\, Acc_{k}^{(2)} == 1)
\end{equation} 

where $\overline{Acc}$ is the Balanced Pairs Accuracy, $B$ denotes the number of complementary pairs, $Acc_{k}^{(1)}$ and $Acc_{k}^{(2)}$ are the accuracies for two images in the $k$-th pair, computed using Equation (\ref{eq:acc}). $\delta(\cdot)$ returns 1 if both accuracies are 1, and 0 otherwise.

\subsection{Experimental Settings}

\subsubsection{Training Details}
All experiments described in this study were conducted on a single RTX 3090 GPU with 24GB of memory. For the purpose of maintaining consistency and fairness across all tests, we utilized uniform parameters when evaluating a variety of Text Feature Extraction Models (under the same random seed). Specifically, the following settings were applied to each experiment:

\begin{itemize}
    \item \textbf{Training Duration}: Each model was trained for a total of 100 epochs, incorporating an early stopping mechanism to prevent overfitting. Specifically, training would be terminated if there was no improvement in performance observed over a span of 20 consecutive epochs.
    \item \textbf{Learning Rate}: We initiated the training with a learning rate of 0.001. To manage the learning rate dynamically, an exponential decay strategy was used, formulated as:
    \begin{equation}
        lr_{i+1} = lr_i \times 0.5^{(1/lr\_half)}
    \end{equation}
    
    Here, $lr_{i}$ represents the learning rate at the current epoch, and $lr_{i+1}$ denotes the learning rate for the next epoch. The term $lr\_half$ is set to 50,000, indicating that the learning rate is halved every 50,000 epochs.
    \item \textbf{Batch Size}: The batch size was consistently set to 256 for all models.
    \item \textbf{Object Proposals}: In our experiments, the maximum number of object proposals per image was set to 100.
\end{itemize}

\subsubsection{Compared Methods}
Image features are computed from R-CNN model with ResNet-101 as backbone mentioned in Section \ref{Image Encoder}. Different text models are adopted, including basic one-layer Gated Recurrent Unit (GRU), Text Convolutional Neural Network (TextCNN), two-layer Long Short-Term Memory Network(2-Layer-LSTM), Bidirectional GRU (BiGRU), Bidirectional LSTM (BiLSTM), Self-Attention GRU (SA-GRU), Multi-Head Self-Attention GRU (MHA-GRU), Transformer Encoder (TE) and Transformer Encoder with GRU (TE-GRU). Specific settings are as follows:

\begin{itemize}
    \item \textbf{One-Layer Gated Recurrent Unit (GRU)}: The baseline of our experiments, which is directly based on the work of \cite{zhang2018vqacount}.
    \item \textbf{TextCNN}: Similar to the work of \cite{kim-2014-convolutional} except that incorporates a single convolutional layer with an assortment of kernel sizes, specifically 3, 4, and 5 (referring \cite{wang2018learning}).
    \item \textbf{2-Layer-LSTM, BiGRU, BiLSTM}: Extended from the One-Layer GRU and One-layer LSTM by simply adding bidirectional processing or an extra layer.
    \item \textbf{Self-Attention GRU (SA-GRU) and Multi-head Self-Attention GRU (MHA-GRU)}: The SA-GRU incorporates a self-attention mechanism \cite{vaswani2017attention} into the standard GRU architecture, while the MHA-GRU, an extension of SA-GRU, employs a multi-head self-attention mechanism that divides the input sequence into multiple segments for simultaneous processing. In our experiments, the configuration with 5 attention heads was utilized.
    \item \textbf{Transformer Encoder (TE) and Transformer Encoder GRU (TE-GRU)}: Transformer Encoder (TE) directly utilizes the encoder component of the original Transformer model proposed by \cite{vaswani2017attention}. In the Transformer Encoder GRU (TE-GRU), input embeddings are first processed through Transformer Encoder, and then fed into the GRU layer. Similar to MHA-GRU, different heads were tested, and the best performance was achieved when set to 4 or 5.
\end{itemize}

\subsubsection{ConvGRU Setting}
For our ConvGRU model (Section \ref{CGRU}), we experimented with different kernel sizes (1, 2, 3, and 4) for convolutional layers to see how they affect text feature extraction. Following the idea from \cite{szegedy2015going}, we used multiple kernel sizes to capture text features at various scales. Instead of pooling, we chose to concatenate features from different kernels to avoid losing too much textual information since most input questions are quite short. Our main focus was on multi-scale kernels, specifically combinations of k=2+3 and k=1+2+3, to leverage multiple text feature representations.

\subsection{Results}\label{sec:results}

\begin{table}
 \caption{Comparison of different textual models. ``SA'' and ``MHA'' refer to Self-Attention and Multi-Head Self-Attention respectively, and all configurations are the same except for textual encoders.}
  \centering
  \scalebox{1.1} {
  \begin{tabular}{ccccccc}
    \toprule
    \multirow{2}{*}{\textbf{Method}} & \multicolumn{3}{c}{\textbf{VQA Accuracy(\%)}} & \multicolumn{3}{c}{\textbf{Balanced Pair Accuracy(\%)}} \\
    \cmidrule(lr){2-4} \cmidrule(lr){5-7}
    & Number & Count & All & Number & Count & All \\
    \midrule
    RCNN+GRU \cite{zhang2018vqacount} & 49.40 & 57.13 & 65.42 & 22.98 & 26.47 & 37.26 \\
    RCNN+TextCNN & 47.68 & 55.23 & 64.00 & 21.42 & 24.71 & 35.12 \\
    RCNN+2-Layer-LSTM & 48.23 & 55.74 & 64.98 & 22.34 & 25.81 & 36.76 \\
    RCNN+BiGRU & 48.84 & 56.57 & 65.44 & 22.34 & 25.81 & 37.38 \\
    RCNN+BiLSTM & 48.11 & 55.68 & 65.15 & 21.70 & 25.09 & 36.97 \\
    RCNN+SA-GRU & 46.04 & 53.11 & 64.47 & 18.96 & 21.85 & 35.96 \\
    RCNN+MHA-GRU & 48.71 & 56.25 & 64.92 & 22.65 & 26.11 & 36.94 \\
    RCNN+TE & 46.70 & 53.98 & 63.53 & 20.47 & 23.72 & 35.41 \\
    RCNN+TE-GRU & 48.20 & 55.77 & 64.66 & 21.77 & 25.14 & 36.31 \\
    \bottomrule
  \end{tabular}
  }
  \label{tab:comparemodels}
\end{table}

\begin{table}
 \caption{Performance comparison of ConvGRU configurations with varying kernel sizes (denoted by numbers in parentheses) and the residual connections (indicated by 'Res'). “Head” and “Tail” refer to asymmetric head padding and asymmetric tail padding discussed in Section \ref{CGRU}, respectively.}
  \centering
  \small
  \begin{adjustbox}{width=\textwidth,center}
  \begin{tabular}{ccccccccccc}
    \toprule
    \multirow{2}{*}{\textbf{GRU Setting}} & \multicolumn{5}{c}{\textbf{VQA Accuracy(\%)}} & \multicolumn{5}{c}{\textbf{Balanced Pair Accuracy(\%)}} \\
    \cmidrule(lr){2-6} \cmidrule(lr){7-11}
    & Yes/No & Number & Count & Other & All & Yes/No & Number & Count & other & All \\
    \midrule
    GRU \cite{zhang2018vqacount} & 81.81 & 49.40 & 57.13 & 57.20 & 65.42 & 55.01 & 22.98 & 26.47 & 28.29 & 37.26 \\
    + Conv(1) & 81.76 & 49.07 & 56.84 & 57.14 & 65.33 & 54.71 & 22.64 & 26.20 & 28.25 & 37.09 \\
    + Conv(2) & 81.70 & 49.46 & 57.18 & 57.03 & 65.31 & 54.60 & 23.02 & 26.60 & 28.20 & 37.08 \\
    + Conv(3) & 81.80 & 49.22 & 56.86 & 57.21 & 65.40 & 54.93 & 23.00 & 26.54 & 28.40 & 37.29 \\
    + Conv(4) & 81.53 & 46.40 & 53.61 & 57.11 & 64.88 & 54.64 & 19.45 & 22.44 & 28.17 & 36.59 \\
    + Conv(1+2+3) & 81.85 & 49.11 & 56.84 & 57.27 & 65.43 & 54.78 & 22.95 & 26.47 & 28.22 & 37.15 \\
    + Conv(2+3+Res,Tail) & 81.70 & 49.26 & 56.96 & 57.20 & 65.36 & 54.68 & 23.11 & 26.69 & 28.26 & 37.15 \\
    \midrule
    + Conv(3+Res) & 81.88 & 49.36 & 57.12 & 57.19 & 65.44 & 54.93 & \textbf{23.43} & \textbf{27.03} & 28.10 & 37.21 \\
    + Conv(2+3+Res,Head) & 81.96 & 49.53 & 57.26 & 57.33 & \textbf{65.56} & 55.09 & \textbf{23.36} & \textbf{26.94} & 28.41 & \textbf{37.41} \\
    \bottomrule
  \end{tabular}
  \end{adjustbox}
  \label{tab:CGRUResults}
\end{table}

Table \ref{tab:comparemodels} presents the performance of different Text Feature Extraction Models in terms of VQA Accuracy (\%) and Balanced Pair Accuracy (\%) on VQA-v2 validation set. Our experiments show that using complex models on the text modality doesn't always help with VQA tasks. In fact, models with complex features like Transformer Encoders and attention mechanisms (Self-Attention and Multi-Head Self-Attention) perform worse than simpler ones. This suggests that for VQA tasks, where questions are short and have similar meanings, these complex models good at capturing long-distance dependencies or global features might not always be necessary.

On the contrary, local text features seem to play a more pivotal role in determining model accuracy considering the nature of questions in VQA-v2 dataset. This can be seen from the results shown in Table \ref{tab:CGRUResults}. The convolutional layer and residual unit in ConvGRU tend to capture extra local dependencies within the text data effectively without losing information from initial text representations. When comparing the ConvGRU variants with kernel sizes ranging from 1 to 4 (Conv(1) to Conv(4)), a clear pattern emerges: The Conv(1) model, equally performing a linear projection, offers limited benefits in terms of capturing the nuanced relationships between words in a question. On the other hand, models with larger kernel sizes, like Conv(4), demonstrate a decline in performance, suggesting that excessively large kernels may overlook crucial local textual features. This observation implies that kernels larger than 4 are too broad, potentially diluting the model's focus on the immediate contextual relevance between adjacent words. Conversely, kernel sizes of 2 and 3 strike a balance.

This insight led us to explore multi-scale convolution strategies, combining the strengths of kernel sizes 2 and 3. Our experiments confirm that this multi-scale approach, particularly with asymmetric head-padding (Conv(2+3+Res,Head)), marginally outperforms the equivalent configuration with tail-padding. The slight preference for head-padding may be attributed to its emphasis on the initial segments of questions, which often direct the types of questions.

Although the Conv(3+Res) shows competitive, and in some aspects, superior performance to Conv(2+3+Res,Head), the latter's broader applicability across different text scenarios, as further discussed in Section \ref{Qualitative Results}, positions it as a more versatile model. The Conv(2+3+Res,Head) model not only capitalizes on multi-scale convolution benefits but also enhances the interpretability and adaptability of the model to various question types within the VQA framework. This makes it an optimal choice for addressing the diverse challenges presented by the VQA tasks. 

Also, it needs to point out that relatively incremental improvement by ConvGRU over the baseline GRU model may stem from limitations inherent in the Image Encoder, attention mechanisms, and the visual-textual fusion methods employed. Therefore, advancements in these areas might further unlock the potential of ConvGRU in VQA tasks.

\subsection{Unexpected Shortfalls: Why Transformer Encoder Struggles with VQA Text}
The underperformance of Transformer Encoder in Visual Question Answering (VQA) tasks may initially appear surprising, given their success in broader NLP applications. This discrepancy arises from the unique textual characteristics within VQA, where questions are typically short and semantically similar. Unlike many NLP tasks where global contextual insights are crucial, VQA demands precise extraction of information from concise queries.

Transformer Encoder is good at capturing relationships in extensive text sequences through self-attention mechanisms. However, VQA tasks often involve brief textual prompts seeking specific visual details. This context significantly limits the utility of global contextual analysis, as the essence of VQA questions lies in localized cues rather than extensive narrative contexts. Furthermore, the semantic similarity among VQA questions adds another layer of complexity. Transformer Encoder may not effectively prioritize the subtle distinctions critical in VQA. Thus, their architectural advantages, while transformative for general NLP, may inadvertently obscure the localized, specific cues vital for accurate Visual Question Answering.

In essence, the architectural design of Transformer Encoder, though revolutionary for capturing global textual relationships, aligns less effectively with the requirements of VQA tasks. This misalignment underscores the necessity for approaches that emphasize local feature extraction, suggesting a tailored adaptation of text models to better accommodate the concise and semantically concentrated nature of VQA queries.

\subsection{How Short Can VQA Questions Get?}

\begin{figure}
  \centering
  \includegraphics[width=\textwidth]{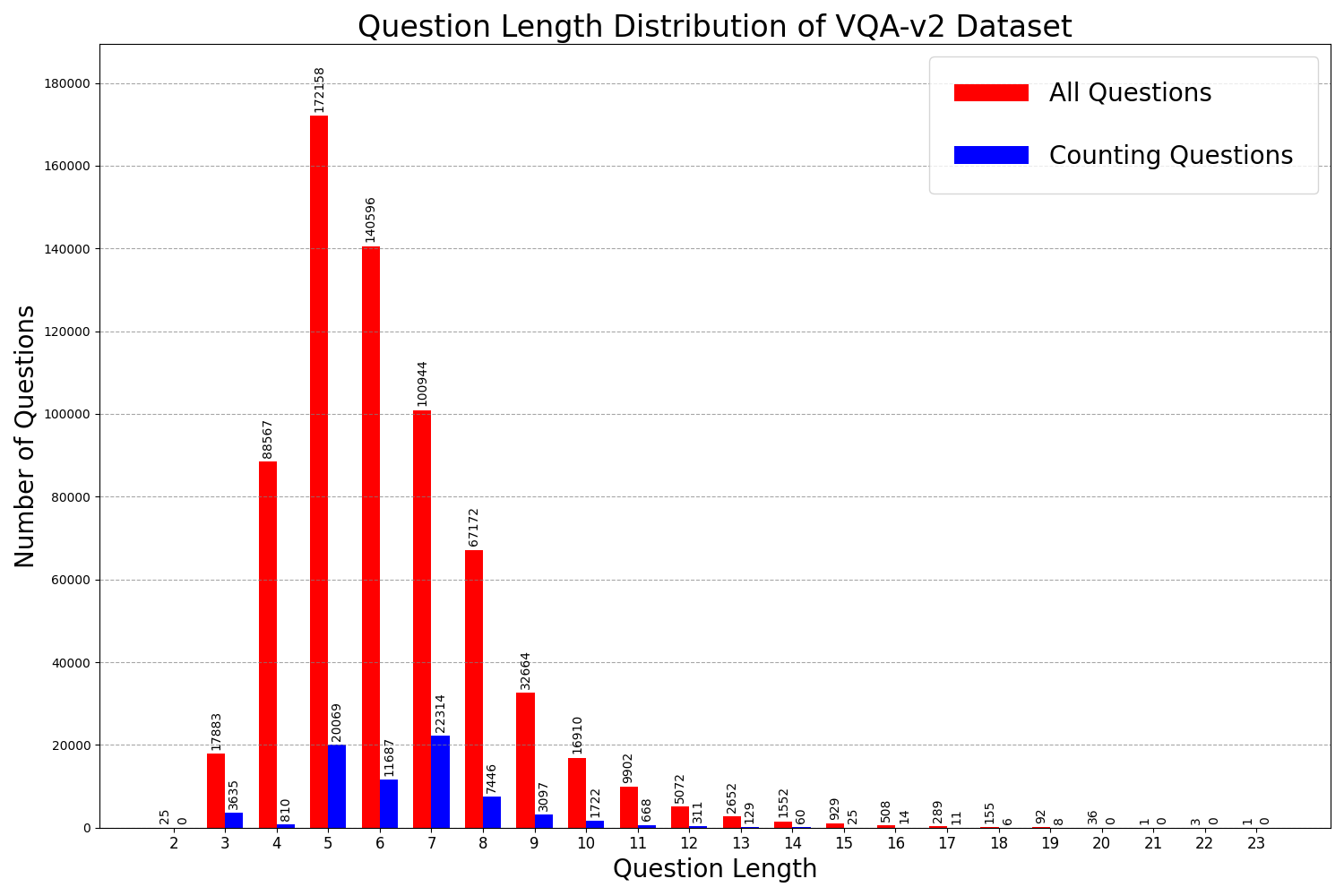}
  \caption{Distribution of question lengths within the VQA-v2 dataset. Red bars represent the total number of questions at each length, while blue bars show the number of counting questions that start with 'How many'.}
  \label{fig:text-length}
\end{figure}

We conduct an analysis of the length distribution of questions in the VQA-v2 dataset to demonstrate how the convolutional extraction of local features can lead to performance improvements.

As shown in Figure \ref{fig:text-length}, we use bar chart to represent the distribution, with the X-axis indicating the length of questions and the Y-axis indicating the frequency of questions at each length. In our comprehensive analysis of 658,111 questions from VQA-v2 dataset, we found that a significant majority, 636,894 questions (96.78\%), have lengths ranging from 3 to 10 words. Notably, 569,437 questions (86.53\%) are within the 4 to 8-word range. A steep decrease in question frequency is observed with increasing length. Focusing on the ``How many'' counting questions (72,012 in total), a predominant 61,516 questions (85.42\%) fall within the 5 to 8-word range, with questions exceeding 10 words represent a mere 4.1\% of counting queries.

These insights highlight the predominance of short questions in VQA-v2 dataset, demonstrating the effectiveness of smaller convolution kernels for local feature extraction. The prevalence of question-type determinants at the start of queries (``How many'', ``Is there'', ``Do'', etc.) further advocates for the superiority of compact kernels. This observation also aligns with the performance dip for k=4 scenarios depicted in Table 3, reinforcing smaller kernels' utility in enhancing textual modality within VQA frameworks.

\subsection{Qualitative Results} \label{Qualitative Results}

\begin{figure}
  \centering
  \includegraphics[width=\textwidth]{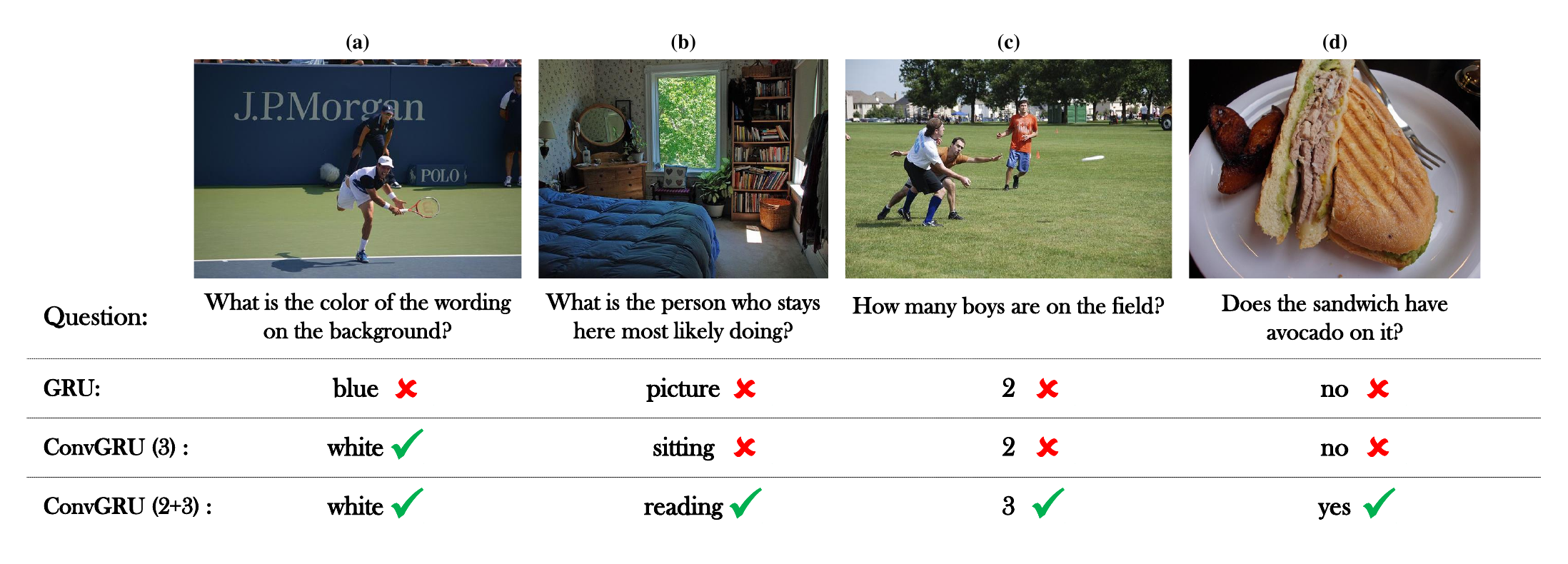}
  \caption{Qualitative results on validation samples indicating how GRU models enhanced with convolutional layer can predict more accurate answers.}
  \label{fig:casestudy}
\end{figure}

In this section, we present four distinct cases from validation set to demonstrate the impact of different GRU configurations on prediction accuracy (Figure \ref{fig:casestudy}). We compare the \textbf{GRU}, \textbf{ConvGRU(3)}, and \textbf{ConvGRU(2+3)} models across cases (a) to (d).

\begin{itemize}
\item As shown in Case (a), the original GRU incorrectly predicts the answer to be ``blue'', likely confusing the color of the background with that of the text. Conversely, both the ConvGRU(3) and the ConvGRU(2+3) models accurately identify the correct answer as ``white''. This illustrates the capability of these enhanced models to capture key phrase pairs such as ``color of'' and ``wording on'' through 2-grams, and richer context through 3-grams like ``the color of'' and ``of the wording,'' which effectively utilize the convolutional layers to discern the finer details of the question.
\item In Case (b), the GRU model gives a completely wrong prediction of ``picture'' as it relies solely on simplistic word-level features. The ConvGRU(3) model improves by predicting ``sitting''—a rationally deduced yet contextually incorrect response, revealing that 3-gram features effectively capture a person’s state or action in a specific location. The ConvGRU(2+3), by incorporating both 2-gram and 3-gram features, correctly comprehends the query's intent and the subject's context, leading to an accurate ``reading'' prediction. This case exemplifies how combining different n-grams can enrich linguistic feature understanding and enhance alignment with image features for more accurate predictions.
\item In both the counting task and the yes/no task, as shown in Case (c) and Case (d), ConvGRU(2+3) showcases best performance. The ability to understand relational context and specific details allows it to correctly count ``3'' boys on the field and accurately confirm the presence of avocado on the sandwich with a ``yes''. These outcomes illustrate the effectiveness of integrating multi-gram features into GRUs.
\end{itemize}

Across all cases, the results consistently demonstrate that ConvGRU(2+3) achieves higher accuracy compared to ConvGRU(3) and the original GRU. This highlights the effectiveness of integrating multi-gram features in enhancing the GRU's comprehension of textual information, thereby improving VQA accuracy without significantly increasing computational costs.

\section{Conclusion}
This paper explored the effectiveness of complex versus simpler models for capturing textual features in Visual Question Answering (VQA) tasks, specifically addressing whether complex sequential models are optimal for this purpose. Our experiments demonstrated that the ConvGRU model, designed to capture local text features, marginally outperformed the baseline GRU model, while advanced models like Self-Attention and Transformer Encoders showed reduced effectiveness. The primary advantage of ConvGRU lies in its ability to balance model performance with efficiency.

Nonetheless, our model is limited in scope, as it does not address other components of the VQA framework, such as the image encoder, where more advanced models beyond the Region-based Convolutional Neural Networks (R-CNN) could potentially improve results. Future research should focus on targeted improvements within both individual components and fusion methods to further optimize multi-modal understanding. Potential directions include exploring advanced neural architectures tailored for visual data and developing fusion techniques that can more effectively integrate visual and textual modalities to foster a deeper and more contextually relevant understanding in VQA tasks.

\bibliographystyle{unsrt}  
\bibliography{main}  

\begin{thebibliography}{10}

\bibitem{antol2015vqa}
Stanislaw Antol, Aishwarya Agrawal, Jiasen Lu, Margaret Mitchell, Dhruv Batra, C~Lawrence Zitnick, and Devi Parikh.
\newblock Vqa: Visual question answering.
\newblock In {\em Proceedings of the IEEE international conference on computer vision}, pages 2425--2433, 2015.

\bibitem{du2024comparison}
Shiyan Du, Jiacheng Li, and Masato Noto.
\newblock Comparison and analysis of three mobilenet-based models for wildfire detection.
\newblock {\em Journal of Advances in Information Technology}, 15(4), 2024.

\bibitem{goyal2017making}
Yash Goyal, Tejas Khot, Douglas Summers-Stay, Dhruv Batra, and Devi Parikh.
\newblock Making the v in vqa matter: Elevating the role of image understanding in visual question answering.
\newblock In {\em Proceedings of the IEEE conference on computer vision and pattern recognition}, pages 6904--6913, 2017.

\bibitem{malinowski2015ask}
Mateusz Malinowski, Marcus Rohrbach, and Mario Fritz.
\newblock Ask your neurons: A neural-based approach to answering questions about images.
\newblock In {\em Proceedings of the IEEE international conference on computer vision}, pages 1--9, 2015.

\bibitem{wang2018learning}
Zhengyang Wang and Shuiwang Ji.
\newblock Learning convolutional text representations for visual question answering.
\newblock In {\em Proceedings of the 2018 SIAM International Conference on Data Mining}, pages 594--602. SIAM, 2018.

\bibitem{liu2018ivqa}
Feng Liu, Tao Xiang, Timothy~M Hospedales, Wankou Yang, and Changyin Sun.
\newblock ivqa: Inverse visual question answering.
\newblock In {\em Proceedings of the IEEE Conference on Computer Vision and Pattern Recognition}, pages 8611--8619, 2018.

\bibitem{lu2019vilbert}
Jiasen Lu, Dhruv Batra, Devi Parikh, and Stefan Lee.
\newblock Vilbert: Pretraining task-agnostic visiolinguistic representations for vision-and-language tasks.
\newblock {\em Advances in neural information processing systems}, 32, 2019.

\bibitem{li2019visualbert}
Liunian~Harold Li, Mark Yatskar, Da~Yin, Cho-Jui Hsieh, and Kai-Wei Chang.
\newblock Visualbert: A simple and performant baseline for vision and language.
\newblock {\em arXiv preprint arXiv:1908.03557}, 2019.

\bibitem{radford2021learning}
Alec Radford, Jong~Wook Kim, Chris Hallacy, Aditya Ramesh, Gabriel Goh, Sandhini Agarwal, Girish Sastry, Amanda Askell, Pamela Mishkin, Jack Clark, et~al.
\newblock Learning transferable visual models from natural language supervision.
\newblock In {\em International conference on machine learning}, pages 8748--8763. PMLR, 2021.

\bibitem{garderes2020conceptbert}
Fran{\c{c}}ois Gard{\`e}res, Maryam Ziaeefard, Baptiste Abeloos, and Freddy Lecue.
\newblock Conceptbert: Concept-aware representation for visual question answering.
\newblock In {\em Findings of the Association for Computational Linguistics: EMNLP 2020}, pages 489--498, 2020.

\bibitem{ravi2023vlc}
Sahithya Ravi, Aditya Chinchure, Leonid Sigal, Renjie Liao, and Vered Shwartz.
\newblock Vlc-bert: Visual question answering with contextualized commonsense knowledge.
\newblock In {\em Proceedings of the IEEE/CVF Winter Conference on Applications of Computer Vision}, pages 1155--1165, 2023.

\bibitem{song2022clip}
Haoyu Song, Li~Dong, Wei-Nan Zhang, Ting Liu, and Furu Wei.
\newblock Clip models are few-shot learners: Empirical studies on vqa and visual entailment.
\newblock {\em arXiv preprint arXiv:2203.07190}, 2022.

\bibitem{zhou2020unified}
Luowei Zhou, Hamid Palangi, Lei Zhang, Houdong Hu, Jason Corso, and Jianfeng Gao.
\newblock Unified vision-language pre-training for image captioning and vqa.
\newblock In {\em Proceedings of the AAAI conference on artificial intelligence}, volume~34, pages 13041--13049, 2020.

\bibitem{yang2016stacked}
Zichao Yang, Xiaodong He, Jianfeng Gao, Li~Deng, and Alex Smola.
\newblock Stacked attention networks for image question answering.
\newblock In {\em Proceedings of the IEEE conference on computer vision and pattern recognition}, pages 21--29, 2016.

\bibitem{xu2016ask}
Huijuan Xu and Kate Saenko.
\newblock Ask, attend and answer: Exploring question-guided spatial attention for visual question answering.
\newblock In {\em Computer Vision--ECCV 2016: 14th European Conference, Amsterdam, the Netherlands, October 11--14, 2016, Proceedings, Part VII 14}, pages 451--466. Springer, 2016.

\bibitem{seo2016bidirectional}
Minjoon Seo, Aniruddha Kembhavi, Ali Farhadi, and Hannaneh Hajishirzi.
\newblock Bidirectional attention flow for machine comprehension.
\newblock {\em arXiv preprint arXiv:1611.01603}, 2016.

\bibitem{kazemi2017show}
Vahid Kazemi and Ali Elqursh.
\newblock Show, ask, attend, and answer: A strong baseline for visual question answering.
\newblock {\em arXiv preprint arXiv:1704.03162}, 2017.

\bibitem{kim-2014-convolutional}
Yoon Kim.
\newblock Convolutional neural networks for sentence classification.
\newblock In Alessandro Moschitti, Bo~Pang, and Walter Daelemans, editors, {\em Proceedings of the 2014 Conference on Empirical Methods in Natural Language Processing ({EMNLP})}, pages 1746--1751, Doha, Qatar, October 2014. Association for Computational Linguistics.

\bibitem{vaswani2017attention}
Ashish Vaswani, Noam Shazeer, Niki Parmar, Jakob Uszkoreit, Llion Jones, Aidan~N Gomez, {\L}ukasz Kaiser, and Illia Polosukhin.
\newblock Attention is all you need.
\newblock {\em Advances in neural information processing systems}, 30, 2017.

\bibitem{devlin2018bert}
Jacob Devlin, Ming-Wei Chang, Kenton Lee, and Kristina Toutanova.
\newblock Bert: Pre-training of deep bidirectional transformers for language understanding.
\newblock {\em arXiv preprint arXiv:1810.04805}, 2018.

\bibitem{yang2021tap}
Zhengyuan Yang, Yijuan Lu, Jianfeng Wang, Xi~Yin, Dinei Florencio, Lijuan Wang, Cha Zhang, Lei Zhang, and Jiebo Luo.
\newblock Tap: Text-aware pre-training for text-vqa and text-caption.
\newblock In {\em Proceedings of the IEEE/CVF conference on computer vision and pattern recognition}, pages 8751--8761, 2021.

\bibitem{zhang2018vqacount}
Yan Zhang, Jonathon Hare, and Adam Pr\"ugel-Bennett.
\newblock Learning to count objects in natural images for visual question answering.
\newblock In {\em International Conference on Learning Representations}, 2018.

\bibitem{he2016deep}
Kaiming He, Xiangyu Zhang, Shaoqing Ren, and Jian Sun.
\newblock Deep residual learning for image recognition.
\newblock In {\em Proceedings of the IEEE conference on computer vision and pattern recognition}, pages 770--778, 2016.

\bibitem{wang2016semantic}
Peng Wang, Bo~Xu, Jiaming Xu, Guanhua Tian, Cheng-Lin Liu, and Hongwei Hao.
\newblock Semantic expansion using word embedding clustering and convolutional neural network for improving short text classification.
\newblock {\em Neurocomputing}, 174:806--814, 2016.

\bibitem{wu2019convolution}
Shuang Wu, Guanrui Wang, Pei Tang, Feng Chen, and Luping Shi.
\newblock Convolution with even-sized kernels and symmetric padding.
\newblock {\em Advances in Neural Information Processing Systems}, 32, 2019.

\bibitem{teney2018tips}
Damien Teney, Peter Anderson, Xiaodong He, and Anton Van Den~Hengel.
\newblock Tips and tricks for visual question answering: Learnings from the 2017 challenge.
\newblock In {\em Proceedings of the IEEE conference on computer vision and pattern recognition}, pages 4223--4232, 2018.

\bibitem{szegedy2015going}
Christian Szegedy, Wei Liu, Yangqing Jia, Pierre Sermanet, Scott Reed, Dragomir Anguelov, Dumitru Erhan, Vincent Vanhoucke, and Andrew Rabinovich.
\newblock Going deeper with convolutions.
\newblock In {\em Proceedings of the IEEE conference on computer vision and pattern recognition}, pages 1--9, 2015.

\end{thebibliography}

\end{document}